\pdfoutput=1

\documentclass[11pt]{article}

\usepackage{ACL2023}

\usepackage{times}
\usepackage{latexsym}

\usepackage[T1]{fontenc}

\usepackage[utf8]{inputenc}

\usepackage{microtype}

\usepackage{array}
\usepackage{enumitem}
\usepackage{float}
\usepackage{graphicx}
\usepackage{setspace}
\usepackage{bm}
\usepackage{amsfonts}
\usepackage{amssymb}
\usepackage{booktabs}
\usepackage{multirow}

\usepackage{soul}
\graphicspath{ {./} }
\usepackage{url}
\usepackage{xcolor}

\usepackage{amsmath}

\usepackage{algorithm}
\usepackage{algorithmic}
\usepackage{float}  
\usepackage{lipsum}
\usepackage{hyperref}

\definecolor{ao}{rgb}{0.0, 0.5, 0.0}
\newcommand{\vic}[1]{\textcolor{black}{#1}}

\newcommand{\jeff}[1]{\textcolor{black}{#1}}

\title{BUCA: A Binary Classification Approach to Unsupervised Commonsense Question Answering}

\author{Jie He$^1$  \and Simon Chi Lok U$^1$  \and V\'ictor Guti\'errez-Basulto$^2$  \and Jeff Z. Pan$^1$  \\
         $^1$ ILCC, School of Informatics, University of Edinburgh, UK  \\  $^2$ School of Computer Science and Informatics, Cardiff University, UK
           \\ \tt {j.he@ed.ac.uk, c.l.u@sms.ed.ac.uk}\\ \quad {\tt gutierrezbasultov@cardiff.ac.uk,  j.z.pan@ed.ac.uk}
}

\begin{document}
\maketitle
\begin{abstract}

\vic {Unsupervised commonsense reasoning (UCR) is becoming increasingly popular as the construction of commonsense reasoning datasets is expensive, and they are inevitably limited in their scope. A popular approach to UCR is to fine-tune language models with external knowledge (e.g., knowledge graphs), but this usually requires a large number of training examples. In this paper, we propose to transform the downstream multiple choice question answering task into a simpler binary classification task by ranking all candidate answers according to their reasonableness. To this end, for training the model, we convert the knowledge graph triples into reasonable and unreasonable texts. Extensive experimental results show the effectiveness of our approach on various multiple choice question answering benchmarks. Furthermore, compared with existing UCR approaches using KGs, ours is less data hungry. Our code is  available at \href{https://github.com/probe2/BUCA}{https://github.com/probe2/BUCA}}.

\end{abstract}

\section{Introduction}
\vic{Commonsense reasoning  has recently  received significant  attention in   NLP  research \cite{Bhargava_Ng_2022}, 
with a vast amount of  datasets  now available~\cite{wsc,copa,siqa,event2mind,piqa,commonsenseqa,long-etal-2020-shallow,long-etal-2020-ted}. Most existing methods for commonsense reasoning either fine-tune large language models (LMs) on these  datasets  \cite{rainbow} or use  knowledge graphs (KGs)~\cite{PVGW2017} to train LMs  \cite{k-bert,yasunaga2022dragon}.}
%
However, 
it is not always possible to have relevant training data available, it is thus crucial to develop unsupervised  approaches to commonsense reasoning that do not  rely on  labeled data.

\vic{In this paper, we focus on the unsupervised multiple choice question answering (QA) task: given a question and a set of answer options, the model 
is expected to predict the most likely option. We propose \textbf{BUCA},  a binary classification framework for unsupervised commonsense QA.
Our method roughly works as follows: we first convert knowledge graph triples  into  textual form using manually written templates, and  generate positive and negative question-answer pairs. 
We then fine-tune a pre-trained language model, and leverage contrastive learning to increase the ability to distinguish reasonable from unreasonable ones. Finally, we input each question and all options of the downstream commonsense QA task into  BUCA to obtain the reasonableness scores 
and select the answer with the highest reasonableness score as the predicted answer. Experimental results on various commonsense reasoning benchmarks 
show the effectiveness of our proposed BUCA framework. Our main contributions are:}

\begin{figure}[!tbp]
\centering
\setlength{\belowcaptionskip}{-0.45cm} 
\includegraphics[scale=0.3]{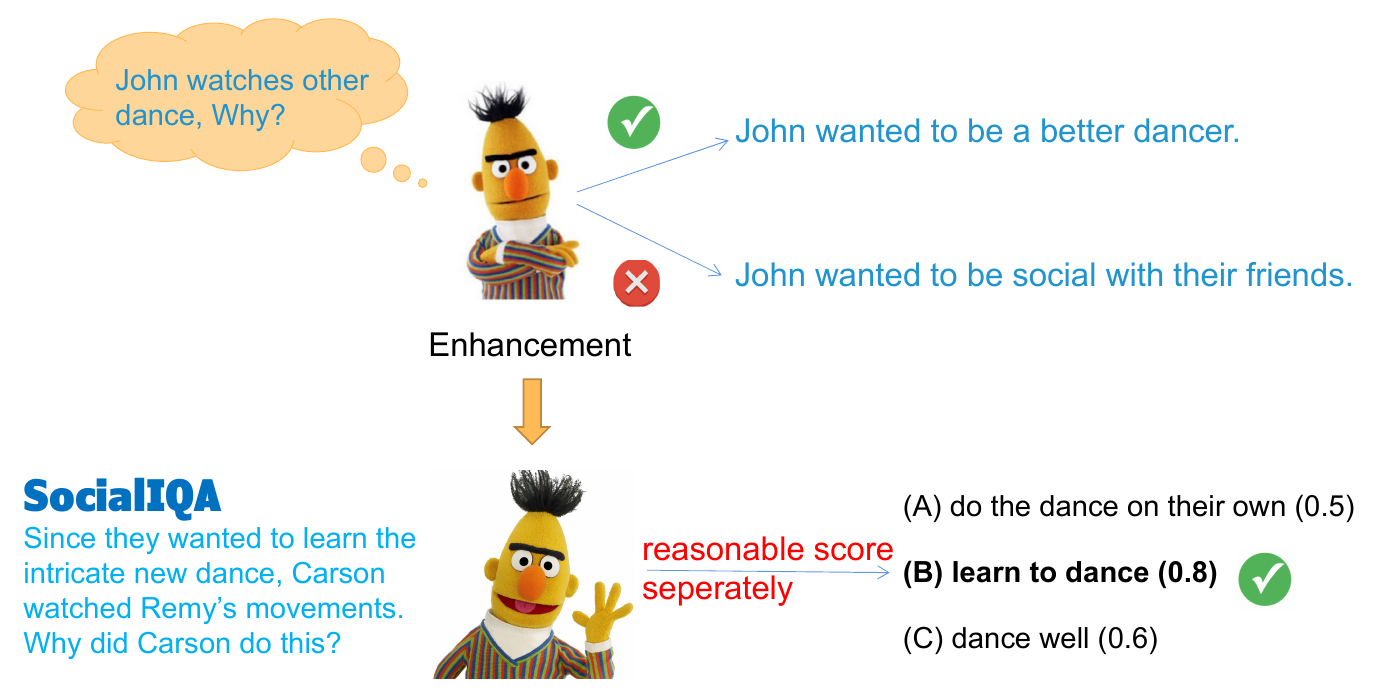}
\caption
{After BUCA is trained on the above question from the training set, it is then able to rate the reasonableness of each sentence of the downstream task.} 
\label{figure_main_model}
\end{figure}
\begin{itemize}
\item 
\vic{We propose a binary classification approach to using  KGs for unsupervised commonsense question answering.} 
\item 
\vic{We conduct extensive experiments, showing the effectiveness of our approach by using much less data.} 
\end{itemize}
\section{Related work}
\vic{Language models are widely used in unsupervised commonsense inference tasks, e.g.  as an additional knowledge source  or as a scoring model. \citet{2019-explain} propose an explanation generation model  for the CommonsenseQA dataset. Self-talk \cite{self-talk} uses prompts to stimulate GPT and generate new knowledge.
SEQA \cite{2021-semantic} generates several candidate answers using GPT2 and then ranks each them. }  

Another research direction in unsupervised commonsense reasoning is  the use  of e.g. commonsense KGs~\cite{SCH2016,RRPP+2019,MBBC2020,he-etal-2022-evaluating} to train the model~\cite{CGCH+2021,GCZC+2023,long-webber-2022-facilitating,long-etal-2024-multi,long2024leveraginghierarchicalprototypesverbalizer,}.
In \citet{baral-2020-self}, given 
\jeff{the inputs of context, question and answer}, the model learns to generate one of the inputs given \jeff{the} other two. 
\citet{Knowledge-driven} \jeff{update} the model with a margin ranking loss computed on positive and negative examples from KGs.  
MICO \cite{mico} uses the distance between the positive and negative question-answer pairs obtained from the KG to calculate the loss. However, 
\jeff{all} of the above approaches \jeff{demand a}   large amount of training data, sometimes reaching million of training \jeff{samples}, while BUCA only needs tens of thousands, cf. Table~\ref{tab:dataset_statistics}.  The most similar to our work is NLI-KB \cite{nli-kb}, which trains a model on NLI data, then applies the corresponding knowledge to each question-answer pair on the downstream task. 
\jeff{Our} paper, \jeff{instead},   \jeff{shows} that is not the NLI data but the retrieved knowledge that \jeff{helps}.

\section{Methodology} \label{sec:met}

\vic{We focus on the following  multiple choice question answering (QA) task: given a question $q$ and a set of options $A$, the model should select the most likely single answer $A_i \in A$. We consider an unsupervised setting in which  the model does not have access to the training  or  validation  data. \jeff{Our} BUCA approach first trains the model  \jeff{with} 
a knowledge graph and then uses the trained model to test on multiple QA downstream tasks. Formally, a \emph{knowledge graph (KG)}~\cite{PVGW2017} $\mathcal G$ is a tuple $(V, R, T)$, where $V$ is a set of entities, $E$ is a set of relation types and $T$ is a set of triples of the form $(h,r,t)$ with $h,t \in V$ the \emph{head} and \emph{tail} entities and $r \in R$ the \jeff{\emph{relation}} of the triple connecting $h$ and $t$.}

\vic{Our approach has three main components: knowledge graph transfer to training data, training loss design, and downstream task testing:}


\paragraph{Converting Triples into Binary Classification Training Data.}

Inspired by previous work~\cite{mico}, each KG triple is converted into  question-answer pairs by using pre-defined templates, so that the obtained pairs are then used as the input of  the classification task.  We  use the templates provided in \cite{hwang_comet-atomic_2020}.
For example,  the ATOMIC triple 
\textit{(PersonX thanks PersonY afterwards, isAfter, PersonX asked PersonY for help on her homework)} can be converted to “\textit{After PersonX asked PersonY for help on her homework, PersonX thanks PersonY afterwards}”.  In the appendix we show the distribution of the converted sequence pairs. Along with the correct QA pairs created from the  KG triples, our framework is also trained on negative QA pairs, so it can better discriminate between reasonable and unreasonable QA pairs. More precisely, in the training dataset, each correct QA pair generated from a triple $\textit{tp}= (h,r,t)$ has a corresponding negative pair obtained from a variation of \textit{tp} in which $t$ is substituted by $t'$, which is randomly drawn from the existing tails in the KG.
\begin{table*}[t]
\footnotesize
    \centering
    \begin{tabular}{cccccccccc}
    \hline
         \multirow{2}{*}{Methods}&\multirow{2}{*}{Backbone}&\multirow{2}{*}{Knowledge Source}&
        \multicolumn{2}{c}{COPA} & \multicolumn{2}{c}{OpenbookQA}&SIQA&CSQA&SCT\\
        &&&dev&test&dev&test&dev&dev&dev\\
        \hline
        Random&-&-&50.0&50.0&25.0&25.0&33.3&25.0&50.0\\
        RoBERTa-L&RoBERTa-L&-&54.8&58.4&31.2&31.6&39.7&31.2&65.0\\
        GPT2-L&GPT2-L&-&62.4&63.6&31.2&29.4&42.8&40.4&66.7\\
        Self-talk&GPT2&GPT2&66.0&-&28.4&30.8&46.2&32.4&-\\
        Dou&ALBERT&ALBERT&-&-&\underline{41.6}&\underline{39.8}&44.1&50.9&-\\
        Wang&GPT2&GPT2&69.8&-&-&-&47.3&-&71.6\\
        \hline
        SMLM&RoBERTa-L&e.g., ATOMIC&-&-&34.6&33.8&48.5&38.8&-\\
        MICO&RoBERTa-L&Concept&73.2&75.2&-&-&44.6&51.0&-\\
        MICO&RoBERTa-L&ATOMIC&\underline{79.4}&\underline{77.4}&-&-&56.0&44.2&-\\
        NLI-KB&RoBERTa-L&Concept&65.0&62.2&35.0&35.6&46.9&49.0&71.2\\
        NLI-KB&RoBERTa-L&ATOMIC&65.2&61.6&39.0&37.2&46.7&52.1&\underline{72.1}\\
         Ma&RoBERTa-L&CSKG&-&-&-&-&\underline{63.2}&\underline{67.4}&-\\
        \hline
        BUCA& RoBERTa-L/TBL&Concept&84.4&\textbf{90.6}&43.0&47.2&53.5&63.5&87.3\\
       BUCA& RoBERTa-L/MRL&Concept&\textbf{86.2}&89.6&45.2&\textbf{47.6}&52.6&\textbf{65.4}&88.0\\
        BUCA&RoBERTa-L/TBL&ATOMIC&85.0&86.0&\textbf{45.8}&44.2&60.2&58.7&\textbf{88.4}\\
       
       BUCA& RoBERTa-L/MRL&ATOMIC&84.6&87.8&43.2&46.0&\textbf{61.4}&60.3&85.5\\
    \hline
    \end{tabular}
    \caption{Accuracy (\%) on five public benchmarks. Our best scores  are highlighted in \textbf{bold}, and the results for the best performing baseline are \underline{underlined}. Recall that TBL and MRL refer to the loss functions used in  BUCA. 
    }
    \label{tab:model_results}
\end{table*}

\paragraph{Training Loss.}
\vic{For our binary classification model, we add a classification head with two nodes to the pre-trained language model. After normalizing the values on these two nodes, we can obtain reasonable and unreasonable scores for the QA pairs. From the triple conversion step, we  obtained $n$ training examples, each consisting of a question $q$, correct answer $a_c$, and incorrect answer $a_w$. For each question-answer pair, we can  then obtain the reasonable and unreasonable scores $r_{i}^{+}$ and $r_{i}^{-}$ after applying a softmax layer. In each loss calculation, we jointly consider the correct and incorrect answers. For binary classification, we use two kinds of losses: \emph{Traditional Binary Loss (TBL).}}  

\begin{equation*}
     \mathcal{L}=-\sum_{i=1}^{n}({log(p_{a_c}^{+})+log(p_{a_w}^{-})})
\end{equation*}
\vic{where $p_{a_c}^{+}$ and $p_{a_w}^{-}$ are  the probabilities of correct and incorrect answers, respectively corresponding to reasonable and unreasonable scores.}

 \noindent \textit{Margin Ranking Loss.}
\begin{equation*}
\begin{split}
     \mathcal{L}=\sum_{i=1}^{n}max(0,\eta-{log(p_{a_c}^{+})+log(p_{a_w}^{+})})\\ +max(0,\eta-{log(p_{a_w}^{-})+log(p_{a_c}^{-})})
\end{split}
\end{equation*}
\vic{where $\eta$ is a margin threshold hyper-parameter.}

\vic{In order to pull the representational distance between reasonable question-answer pairs as close as possible and to push the representational distance between reasonable and unreasonable ones as far as possible, we use supervised contrastive learning \cite{2021supervised} along with the binary classification. This is done by considering as positive examples of a given example within a category, all those examples within the same category. }

\noindent \textit{Contrastive Loss of  the $i$-th QA pair}
\begin{equation*}
    \mathcal{L}_{scl}=\sum_{j=1}^{N}1_{y_i=y_j} log\frac{e^{sim(h_j,h_i)\tau}}{\sum_{k=1}^{N}1_{i\ne k}e^{sim(h_k,h_i)/\tau}}
  \label{eq3}
\end{equation*}
where $\tau$ is the temperature
parameter and $h$ denotes the feature vector.

\paragraph{Inference.}
\vic{In the prediction phase for each candidate answer, we calculate its reasonableness score. We choose the answer with the highest reasonableness  score as the predicted answer.} 


\section{Experiments}
In this section, we first describe our experiments on five commonsense question answering datasets, followed by ablation studies and data analysis.


\subsection{Datasets and Baselines}
\vic{We use two well-known commonsense KGs for training our framework: ConceptNet~\cite{concept} and ATOMIC~\cite{sap_atomic_2018}. For evaluation, we use five commonsense QA datasets: COPA~\cite{copa}, OpenBookQA~\cite{openbook}, SIQA~\cite{siqa}, CSQA~\cite{commonsenseqa}, and SCT~\cite{sct}, covering a wide range of topics within commonsense reasoning.} 
%
\vic{We compare our approach with various baselines: RoBERTa-Large~\cite{roberta}, GPT2 \cite{radford2019language}, Self-talk \cite{self-talk}, Dou \cite{dou2022improving}, Wang \cite{zhao-2022-art} and other unsupervised systems using KGs: SMLM \cite{baral-2020-self}, MICO \cite{mico}, NLI-KB \cite{nli-kb} and Ma \cite{Knowledge-driven}. Most reported results are collected from the literature.   For NLI-KB, we used   their publicly available code to get the results. } 

\vic{Details of the KGs and datasets, as well as implementation details, can be found in the appendix. }
\begin{table}[h!]
\footnotesize
    \centering
    \begin{tabular}{ccccccc}
    \hline
        Methods&Dataset& Train Pair& Valid Pair\\
        \hline 
        Ma&ConceptNet&363,646&19,139\\
        Ma&ATOMIC&534,834&60,289\\
        Ma&WikiData&42,342&2,229\\
        Ma&WordNet&256,922&13,523\\
        MICO&ConceptNet&163,840&19,592\\
        MICO&ATOMIC&1,221,072&48,710\\
        \hline
        BUCA&ConceptNet&65,536&7,836\\
        BUCA&ATOMIC&61,053&2,435\\
    \hline
    \end{tabular}
    \caption{Statistics for the training and validation data used by Ma, MICO and BUCA.}
    \label{tab:dataset_statistics}
\end{table}

\begin{table*}[]
\small
    \centering
    \begin{tabular}{lcccccccc}
    \hline
         \multirow{2}{*}{Backbone}&\multirow{2}{*}{CKG}&
        \multicolumn{2}{c}{COPA} & \multicolumn{2}{c}{OpenbookQA}&SIQA&CSQA&SCT\\
        &&dev&test&dev&test&dev&dev&dev\\
        \hline 
        BERT-base&Concept&63.0&67.6&29.6&32.8&40.5&49.6&64.9\\
        BERT-base&ATOMIC&64.8&73.2&31.2&34.0&45.0&45.3&68.7\\
        \hline
        RoBERTa-base&Concept&70.0&72.8&30.0&32.8&46.6&49.0&65.6\\
        RoBERTa-base&ATOMIC&70.4&77.4&33.4&34.2&50.6&46.9&70.6\\
        \hline
       RoBERTa-large&Concept&86.2&89.6&45.2&47.6&52.6&65.4&88.0\\
       RoBERTa-large&ATOMIC&84.6&87.8&43.2&46.0&61.4&60.3&85.5\\
        
    \hline
    \end{tabular}
    \caption{Backbone model study}
    \label{tab:Backbone}
\end{table*}
\begin{table*}[]
\small
    \centering
    \begin{tabular}{lcccccccc}
    \hline
         \multirow{2}{*}{Backbone}&\multirow{2}{*}{CKG}&
        \multicolumn{2}{c}{COPA} & \multicolumn{2}{c}{OpenbookQA}&SIQA&CSQA&SCT\\
        &&dev&test&dev&test&dev&dev&dev\\
        \hline 
       RoBERTa-large&Concept&86.2&89.6&45.2&47.6&52.6&65.4&88.0\\
       w/o contrastive&Concept&83.3&89.0&42.6&46.8&51.9&64.5&87.0\\
       RoBERTa-large&ATOMIC&84.6&87.8&43.2&46.0&61.4&60.3&85.5\\
       w/o contrastive&ATOMIC&84.2&86.6&42.0&44.0&60.6&59.8&84.1\\
    \hline
    \end{tabular}
    \caption{The influence of contrastive learning}
    \label{tab:contrastive_learning}
\end{table*}

\subsection{Main results}
Table \ref{tab:model_results}  shows the results for the five benchmarks. Overall, BUCA achieves the best performance  on all datasets. More precisely, our results respectively outperform baselines on the validation and test sets as follows:
MICO by 6.8\% and 13.2\% on  COPA; Dou by 4.2\% and 7.8\% on OpenbookQA. We also outperform  MICO by 5.4\% on SIQA; NLI-KB by 13.3\%  on CSQA, and  NLI-KB by 16.3\% on SCT. Ma does not provide results for COPA, OpenBookQA and SCT, but it achieves state-of-the-art results   on CSQA 67.4 and on SIQA  63.2, while BUCA's best results respectively are 65.4 and 61.4. However, Ma uses multiple KGs to train a single model, ConceptNet, WordNet, and Wikidata for CSQA and ATOMIC, ConceptNet, WordNet, and Wikidata for SIQA,  with a total training data of 662,909 and 1,197,742, while  BUCA only uses 65,536 and 61,530, cf. Table~\ref{tab:dataset_statistics}. Considering the difference on used training data and the closeness of results, BUCA's approach  clearly demonstrates its effectiveness.
We can also observe the same trend as  in MICO:  ConceptNet is more helpful for CSQA and ATOMIC is more helpful for SIQA. This is explained by the fact that  SIQA is built based on ATOMIC and CSQA is built based on ConceptNet. On other datasets our framework shows similiar behavior with both KGs. As for the loss functions, the margin ranking loss is on average 0.8\% higher than the binary loss on ConceptNet, and 0.1\% higher  on  ATOMIC.
These results are explained by the fact that the ranking loss separates more the scores between reasonable and unreasonable answers.
In light of this, we will only consider margin ranking loss in the below analysis.


\subsection{Ablation Studies} \label{sec:ablation}

\vic{In this section, we analyze the effects of the backbone models, the effect of contrastive learning, and explore the vocabulary overlap between the knowledge training set and the downstream task as well as the accuracy of our BUCA method. }

        

\paragraph{Backbone Pre-trained LMs}
\vic{Our experiments using different backbone models show  that in general the stronger the PLM  the better the performance on the downstream task.   Regarding the KGs, in the BERT-base and RoBERTa-base variants, the ATOMIC-trained models perform better than the ConceptNet-trained models, while in the RoBERTa-large one they perform similarly. This might be explained by the fact  that as the model capacity increases it has more inherently available event-like commonsense knowledge, necessary in the ATOMIC-based datasets. Detailed results are shown in Table~\ref{tab:Backbone}.
}


\paragraph{Effects of Contrastive Learning}
\vic{Our experiments show that the RoBERTa-large variant with contrastive learning outperforms the version without it on all datasets, regardless of the used KG.  Detailed results are shown in Table \ref{tab:contrastive_learning}.}  



\paragraph{Accuracy of the Binary Classifier}

\vic{Inspired by \citet{team}, we evaluate how often input sequences corresponding to correct and incorrect answers are accurately predicted. To this end, we use the RoBERTa-large variant trained on ATOMIC. Table \ref{tab:acc} shows that our model tends to predict all   answers as reasonable since in our training set the  negative examples are randomly selected,     many QA pairs are semantically irrelevant or even ungrammatical. For the manually crafted candidate answers, many of them are semantically relevant and grammatical, so our model predicts them  as reasonable. We   also see that the accuracy metrics for  SCT and COPA  are the highest. Our findings are consistent with \citet{team}. 
 }



\begin{table*}[]
\small
    \centering
    \begin{tabular}{lccccc}
    \hline
         \multirow{2}{*}{Dataset}& \multicolumn{5}{c}{Prediction All}\\
        &Neg&Pos&Incor as Neg&Cor as Pos&Accurate\\
        \hline 
COPA (dev)&0.2&88.0&11.2&99.0&11.0\\
COPA (test)&0.4&88.4&11.2&99.2&10.8\\
OpenbookQA (dev)&1.4&67.8&4.8&93.2&3.4\\
OpenbookQA (test)&1.8&73.8&2.8&93.0&1.0\\
SIQA (dev)&6.3&50.2&15.7&86.7&9.4\\
CSQA (dev)&1.2&35.1&6.5&94.2&5.2\\
SCT (dev)&0.3&87.8&11.8&99.4&11.6\\
    \hline
    \end{tabular}
    \caption{The Neg and Pos column indicate \% of instances for which all answer choices are predicted as negative or positive. The Incor as Neg, Cor as Pos, and Accurate column indicate \% of instances for which all incorrect answers are predicted as negative, the correct answer is predicted as positive, and all answers are predicted accurately as negative or positive. Accurate is the intersection of Incor as Neg and Cor as Pos.}
    \label{tab:acc}
\end{table*}
\subsection{Data Analysis} \label{sec:ana}

\vic{To better understand why transfer learning from CKGs is more suitable than from other datasets (i.e. MNLI or QNLI) in the  commonsense QA task, we performed an analysis on the training data  in \textbf{NLI-KB} \cite{nli-kb} and the used CKGs. Following \cite{word_overlap},  we first compare the vocabulary overlap of ConceptNet, ATOMIC and MNLI (training data) with our evaluation QA datasets. We follow the definition of overlap introduced in~\cite{word_overlap}.} 
\vic{Table \ref{tab:vocab_overlap} shows that MNLI has higher vocabulary overlap with all the evaluation datasets than both used CKGs. However, the results for NLI-KB in Table \ref{tab:model_results} show that the vocabulary overlap is not a key factor for performance as otherwise,
NLI-KB fine-tuned with the NLI datasets (before injecting knowledge) should perform  better that the other models in the downstream task due to the high lexical similarity. }
\begin{table}[!th]
\small
    \centering
    \begin{tabular}{lccc}
    \hline
         
        &Concept&ATOMIC&MNLI\\
        \hline 
COPA (dev)&50.4&70.0&98.0\\
COPA (test)&52.1&71.9&86.4\\
OpenbookQA (dev)&48.4&54.8&92.1\\
OpenbookQA (test)&48.8&55.2&93.1\\
SIQA (dev)&37.3&54.6&94.5\\
CSQA (dev)&59.1&63.2&85.0\\
SCT (dev)&41.2&57.5&94.5\\
    \hline
    \end{tabular}
    \caption{Vocabulary Overlap}
    \label{tab:vocab_overlap}

\end{table}

\begin{table}[H]
\renewcommand*{\arraystretch}{1.4}

\scriptsize
    \centering
    \begin{tabular}{cp{4cm}}
    
    \hline
         \multirow{4}{*}{SIQA Example}&\textbf{Question}: After a long grueling semester, Tracy took the final exam and finished their course today.  Now they would graduate. Why did Tracy do this? 
         
         \textbf{Answer}: \textit{complete their degree on time}\\
        \hline
        \multirow{2}{*}{MNLI}&Because I had a deadline.  \textbf{This entails I had to finish by that time.}\\ \cline{2-2}

        \hline
        \noalign{\vskip 0.00in}

        \multirow{2}{*}{ATOMIC}&\textbf{Tracy wants finish before time expires.} because Tracy takes the exam. \\\cline{2-2}
    \hline
                    \noalign{\vskip 0.00in}
    \multirow{1}{*}{ConceptNet}&pass class causes \textbf{graduation.}\\\cline{2-2}
                    \noalign{\vskip 0.00in}
    \hline
    \end{tabular}
    \caption{Alternative answers for SIQA-question.}
    \label{tab:mnli}
\end{table}

\vic{We also analyze the distance to the sentence embeddings. Our results show that the MNLI entries performed poorly in commonsense knowledge retrieval for SIQA-queries as they are not reasonable answers.  In contrast, the sentences generated from ATOMIC and ConceptNet successfully pair the SIQA-questions with reasonable answers. This reveals that, although MNLI has a higher lexical coverage, MNLI does not have suitable examples to match SIQA questions. Thus models fine-tuned with the NLI dataset hardly get any benefit for downstream commonsense reasoning tasks. Tables \ref{tab:mnli} and \ref{tab:csqa_question_main}  present a random sample showing this, where reasonable alternatives are in bold. 
}
\begin{table}[H]
\renewcommand*{\arraystretch}{1.4}

\scriptsize
    \centering
    \begin{tabular}{cp{4cm}}

    \hline
         \multirow{2}{*}{CSQA Example}&\textbf{Question}: If you have leftover cake, where would you put it?
         
         \textbf{Answer}: \textit{refrigerator}\\
        \hline
        \multirow{5}{*}{MNLI}&In the waste-paper basket.  \textbf{This entails in the garbage bin.}\\ \cline{2-2}

       &In the middle of the dinner plate (or is it a base drum?) \textbf{This entails in the center of the dinner plate.}\\\cline{2-2}
        &We always keep it in the hall drawer.  \textbf{This entails it's always kept in the drawer in the hall.}  \\\cline{2-2}
        \hline
        \noalign{\vskip 0.00in}

        \multirow{3}{*}{ATOMIC}&John cuts the cake. as a result, \textbf{John wants put the rest of the cake in fridge}\\\cline{2-2}
        &\textbf{John places  in the oven.} but before, John needed to mix the cake ingredients \\\cline{2-2}
         &\textbf{John puts in the fridge.} but before, John needed to grab it off the table \\\cline{2-2}
    \hline
                    \noalign{\vskip 0.00in}
    \multirow{3}{*}{ConceptNet}&\textbf{oven} is the position of cake\\\cline{2-2}
        &\textbf{refrigerator} is the position of moldy leftover\\\cline{2-2}
        &\textbf{fridge} is the position of leftover\\\cline{2-2}
    \hline
    \end{tabular}
    \caption{Alternative answers  for CSQA question.}
    \label{tab:csqa_question_main}
\end{table}



\section{Conclusion}
\vic{We presented a framework converting KGs into positive/negative question-answer pairs to train a binary classification model, discriminating  whether a sentence is reasonable. Extensive experiments show the effectiveness of our approach, while using a reasonably small amount of data. For future work, we will explore how to better select negative cases.}





\section*{Limitations}
\vic{The method to select negative examples could  be improved, as  randomly selecting negative examples for training might lead to identifying most of  examples in the evaluation datasets as reasonable. Secondly, we did not explore using other number of candidates in the training set, we always use 2 candidate answers for each question.}

\section*{Acknowledgments}
This work is supported by  the Chang
Jiang Scholars Program (J2019032).



\bibliography{acl_camera}
\bibliographystyle{acl_natbib}

\clearpage
\appendix \label{sec:appendix}
\noindent \textbf{\large Appendix}





\section{KGs, Datasets, and Implementation}
This section contains more experimental details. In particular, we give details of the used KGs and datasets. We also discuss implementation details. 


\subsubsection*{ConceptNet}
\vic{ConceptNet \cite{concept} is a traditional KG  that focuses on taxonomic, lexical and physical relations (e.g., \textit{IsA}, \textit{RelatedTo}, \textit{PartOf}). In our experiment, we employed the CN-82K version which is uniformly sampled from a larger set of  extracted ConceptNet entity-relations \cite{li_commonsense_2016}.}


\subsubsection*{ATOMIC}
\vic{The ATOMIC KG \cite{sap_atomic_2018} focuses on  social-interaction knowledge about everyday events, and thus has a higher coverage in the field of commonsense query answering. It consists of 880K knowledge triples across 9 relations (e.g. xNeed, oEffect, xReact). This includes mentions of topics such as causes and effects, personal feelings toward actions or events, and conditional statements. The ATOMIC dataset is collected and validated completely through crowdsourcing.}


\smallskip
As seen in Table \ref{tab:dataset_statistics}, in comparison to related works: Ma \cite{Knowledge-driven} and MICO \cite{mico}, our methods used much less data from the CKGs (\textbf{\textasciitilde5-8x} Ma, \textbf{\textasciitilde2-20x} MICO) while still maintaining competitive performance on the evaluation dataset.

\subsection{Generation of QA pairs}

\renewcommand{\arraystretch}{1.05}
\begin{table*}[]
\small
    \centering
    \begin{tabular}{cccl}
    \toprule
        Triple&Source&Negative Triple&Generated QA Pairs\\
       \midrule
       \multirow{3}{*}{(chopstick, AtLocation, table)} & \multirow{3}{*}{ConceptNet} & \multirow{3}{*}{(bread, is created by, flour)} & Q: Chopstick located or found at \\
         {} & {} & {} & \textbf{A: table} \\
         {} & {} & {} & B: flour \\
        \midrule 
       \multirow{2}{*}{(PersonX wants to go to the office, }& \multirow{4}{*}{ATOMIC} & \multirow{2}{*}{(PersonX leaves the room,}
 & Q: PersonX wants to go to the\\
       {} & & {} & \hspace{3mm} office, as a result, PersonX will \\
       oEffect, get dressed up) & {} & xWant, to go somewhere else) & \textbf{A: get dressed up} \\
       {} & {} & {} & B: to go somewhere else \\
       
    \bottomrule
    \end{tabular}
    \caption{QA pairs generated by KG Triples}
    \label{tab:triple_samples}
\end{table*}
The QA pairs were generated using the templates in the ATOMIC paper \cite{hwang_comet-atomic_2020}, which is compatible with relations in both ConceptNet and ATOMIC. These templates help to convert KG triples into natural sentences, examples shown in Table \ref{tab:triple_samples}. The head entity and mapped relation phrases are joined as a question. The correct tail entity and a randomly sampled tail from the dataset are used as the positive and negative answers, respectively, for contrastive learning. 

\subsection{Evaluation Datasets}
\vic{We evaluate our framework using five downstream QA tasks: COPA, OpenBookQA, SIQA, CSQA, and SCT, which covere a wide range of topics within commonsense reasoning. Accuracy is used as the evaluation metric. All experiments are perform in an unsupervised setting, where our model are not train on the source task.}


        

\medskip

\paragraph{Choice of Plausible Alternatives (COPA)} \cite{copa} is a two-choice question-answer dataset designed to evaluate performance in open-domain commonsense causal reasoning. Each entry contains a premise and two possible answers, the task is to select the answers that most likely have a causal relationship with the premise. The dataset consists 500 questions for both debvelopment and test sets.

\paragraph{OpenBookQA} \cite{openbook} is inspired from open book exams that assess human understanding in real life. 
This QA task requires a deeper understanding about both open book facts (e.g., \textit{metals is a heat conductor}) and a broad common knowledge (e.g., \textit{a steal spoon is made of metal}) to answer questions like: \textit{Which of these objects conducts the most heat: A metal spoon, pair of jeans, or cotton made clothing?} It contains 500 multiple-choice science questions for both development and test sets.

\paragraph{SocialIQA (SIQA)} \cite{siqa} contains multiple-choice questions with topics concerned with emotional and social interactions in a variety of everyday situations. Each entry comes with a context, a question, and 3
candidate answers. The questions are generated using the \texttt{ATOMIC} KG by converting triples into question sentences using predefined templates, and
the answers are crowdsourced. The dataset's development split is used as evaluation dataset, containing 1,954 questions.

\paragraph{CommonsenseQA (CSQA)} \cite{commonsenseqa} contains questions focused on various commonsense aspects. Each entry contains a question and five candidate answers. The questions are constructed by crowd workers. The answer candidates include distractors comprised of hand-picked ones or nodes from ConceptNet. The development set is used as evaluation set, containing
1,221 questions.

\paragraph{Story Cloze Test (SCT)} \cite{sct} is a LSDSem’17 shared task, evaluating story understanding and script learning. Each entry contains a four-sentence story and two possible fifth sentences, where the model has to pick the most suitable ending for the story. The development set is used as the evaluation set, containing 1572 different stories.

\subsection{Implementation details}

Our experiments are run on a single A100 GPU card. We  use RoBERTa-Large as our backbone model. The training batch size is 196, and the maximal sequence length for training is 64.  The learning rate is set to 5e-5 for all experiments. For experiments with the margin ranking loss, $\eta$ is set to 1. The validation set is evaluated by accuracy and used  to select a best model for further evaluation. The models are trained for 20 epochs and early stopped when the change of validation loss is within 1\%.

\section{Ablation Studies}


 We present the full results for the ablation studies discussed in Section~\ref{sec:ablation}. Table~\ref{tab:Backbone} for the backbone models study; Table~\ref{tab:contrastive_learning} for the influence of contrastive learning; and Table~\ref{tab:acc} for accuracy.

\section{Data Analysis}
In the  analysis of the distance to sentence embeddings, we treat each entry in the CKG datasets as possible answers and encode them using the SBERT pre-trained model (\textit{all-mpnet-base-v2})~\cite{sentence-bert, sbert-pretrained-model}. Then, the cosine-similarity between the SIQA question and the encoded sentences is calculated to rank their semantic relatedness. 

We retrieved the top 3 answers for each source and listed by similarity score at descending order. Table~\ref{tab:mnli_complete} extends the results presented in Section~\ref{sec:ana};  Table~\ref{tab:copa_question} show the alternative answers from CKG datasets  COPA questions.

\begin{table*}[h!]
\renewcommand*{\arraystretch}{1.4}

\small
    \centering
    \begin{tabular}{cp{10cm}}
    
    \hline
         \multirow{2}{*}{SIQA Example}&\textbf{Question}: After a long grueling semester, Tracy took the final exam and finished their course today.  Now they would graduate. Why did Tracy do this? 
         
         \textbf{Answer}: \textit{complete their degree on time}\\
        \hline
        \multirow{5}{*}{MNLI}&Because I had a deadline. \textbf{This entails I had to finish by that time.}\\ \cline{2-2}

       &The professors went home feeling that history had been made. \textbf{This entails The professors returned home.}\\\cline{2-2}
        &They got married after his first year of law school.\textbf{This entails Their marriage took place after he finished his first year of law school.}\\\cline{2-2}
        \hline
        \noalign{\vskip 0.00in}

        \multirow{3}{*}{ATOMIC}&\textbf{Tracy wants finish before time expires.} because Tracy takes the exam\\\cline{2-2}
        &\textbf{Tracy wanted to get a degree.} as a result Tracy finishes Tracy's test\\\cline{2-2}
         &\textbf{Tracy graduates with a degree.} but before, Tracy needed get pass with good marks.\\\cline{2-2}
    \hline
                    \noalign{\vskip 0.00in}
    \multirow{3}{*}{ConceptNet}&\textbf{pass class causes graduation}\\\cline{2-2}
                    \noalign{\vskip 0.00in}
        &\textbf{study ends} with the event or action graduate\\\cline{2-2}
        &graduation because \textbf{take final exam}\\\cline{2-2}
    \hline
    \end{tabular}
    \caption{Complete results of alternative answers retrieved from MNLI, ATOMIC and ConceptNet for SIQA question. 
    Reasonable alternatives are in bold.}
    \label{tab:mnli_complete}
\end{table*}

    


         



\begin{table*}[!thbp]
\renewcommand*{\arraystretch}{1.4}

\small
    \centering
    \begin{tabular}{cp{10cm}}
    
    \hline
         \multirow{2}{*}{COPA Example}&\textbf{Question}: The boy wanted to be muscular. As a result,
         
         \textbf{Answer}: \textit{He lifted weights.}\\
        \hline
        \multirow{7}{*}{MNLI}&Emboldened, the small boy proceeded. \textbf{This entails the small boy felt bolder and continued.}\\ \cline{2-2}

       &Out of shape, fat boy. \textbf{This entails the boy was obese.}\\\cline{2-2}
        &When Sport Resort won the contract for the construction of a new hotel center for 1200 people around the Olympic Sports Arena (built as a reserve for the future, to have it ready in time for the next championships), Gonzo began to push his weight around, because he felt more secure. \textbf{This entails when Sport Resort won the contract for the construction of a new hotel Gonzo felt more secure.}\\\cline{2-2}
        \hline
        \noalign{\vskip 0.00in}

        \multirow{3}{*}{ATOMIC}&John wanted to build his physique. as a result \textbf{the boy lifts weights}\\\cline{2-2}
        &\textbf{The boy starts working out.} as a result, the boy wants to gain more muscle\\\cline{2-2}
         &\textbf{The boy starts lifting weights.} as a result, the boy will build muscle\\\cline{2-2}
    \hline
                    \noalign{\vskip 0.00in}
    \multirow{3}{*}{ConceptNet}&\textbf{lift could make use of muscle}\\\cline{2-2}
                    \noalign{\vskip 0.00in}
        &\textbf{person desires strong body} \\\cline{2-2}
        &\textbf{build muscle because exercise} \\\cline{2-2}
    \hline
    \end{tabular}
    \caption{Alternative answers from CKGs for COPA question. }
    \label{tab:copa_question}
\end{table*}

\end{document}